\documentclass[letterpaper]{article} 
\usepackage[]{aaai2027}  
\usepackage[hyphens]{url}  
\usepackage{graphicx} 
\usepackage{natbib}  
\usepackage{caption} 
\usepackage{algorithm}
\usepackage{algorithmic}

\usepackage{multirow}

\usepackage{amssymb}
\usepackage{amsmath}

\usepackage{tcolorbox}
\tcbuselibrary{breakable}

\usepackage{makecell}

\usepackage{newfloat}
\usepackage{listings}
\DeclareCaptionStyle{ruled}{labelfont=normalfont,labelsep=colon,strut=off} 
\floatstyle{ruled}
\newfloat{listing}{tb}{lst}{}
\floatname{listing}{Listing}

\usepackage{booktabs}

\nocopyright

\title{OmniTryOn: Video Try-On Anything at Once!}

\author{
    Changliang Xia,
    Chengyou Jia,
    Keshuo Xing,
    Bowen Ping,\\
    Xin Shen,
    Zhuohang Dang,
    Minnan Luo\corresponding
}
\affiliations{
    Xi'an Jiaotong University\\
    Xi'an, China
}

\begin{document}

\maketitle

\begin{abstract}
Although video virtual try-on (VVT) has achieved significant progress, existing methods still exhibit two fundamental limitations: first, they are restricted to single-garment transfer, rendering simultaneous multi-object try-on highly impractical; second, their heavy reliance on explicit external geometric priors (e.g., garment masks) inevitably destroys crucial physical dynamics and degrades visual quality. To bridge this gap, this paper proposes the novel Try-On Anything task, which transfers diverse wearable objects onto a person in a single inference pass. To support and standardize this paradigm, we introduce TryAny-Bench, a comprehensive benchmark encompassing a paired video dataset alongside a tailored evaluation protocol. Furthermore, we present OmniTryOn, a generative framework free of explicit geometric priors. Specifically, OmniTryOn employs a First Frame Wearable Cache strategy, which directly provides diverse wearable objects for the generation process through the initial video frame. To maintain consistency, we propose the Spatiotemporally Consistent RoPE (STC-RoPE), which inherently establishes robust spatiotemporal anchors to strictly preserve complex human motions and background dynamics. Optimized by the proposed Gradual Try-On (GTO) training strategy, our model progressively masters robust multi-object synthesis. Extensive experiments demonstrate that OmniTryOn outperforms specialized VVT models and general video editing baselines on the Try-On Anything. Our code, model, and data is publicly released at \url{https://github.com/xcltql666/OminTryOn}.
\end{abstract}


\section{Introduction}

\begin{figure}[t]
  \centering
  \includegraphics[width=0.8\linewidth]{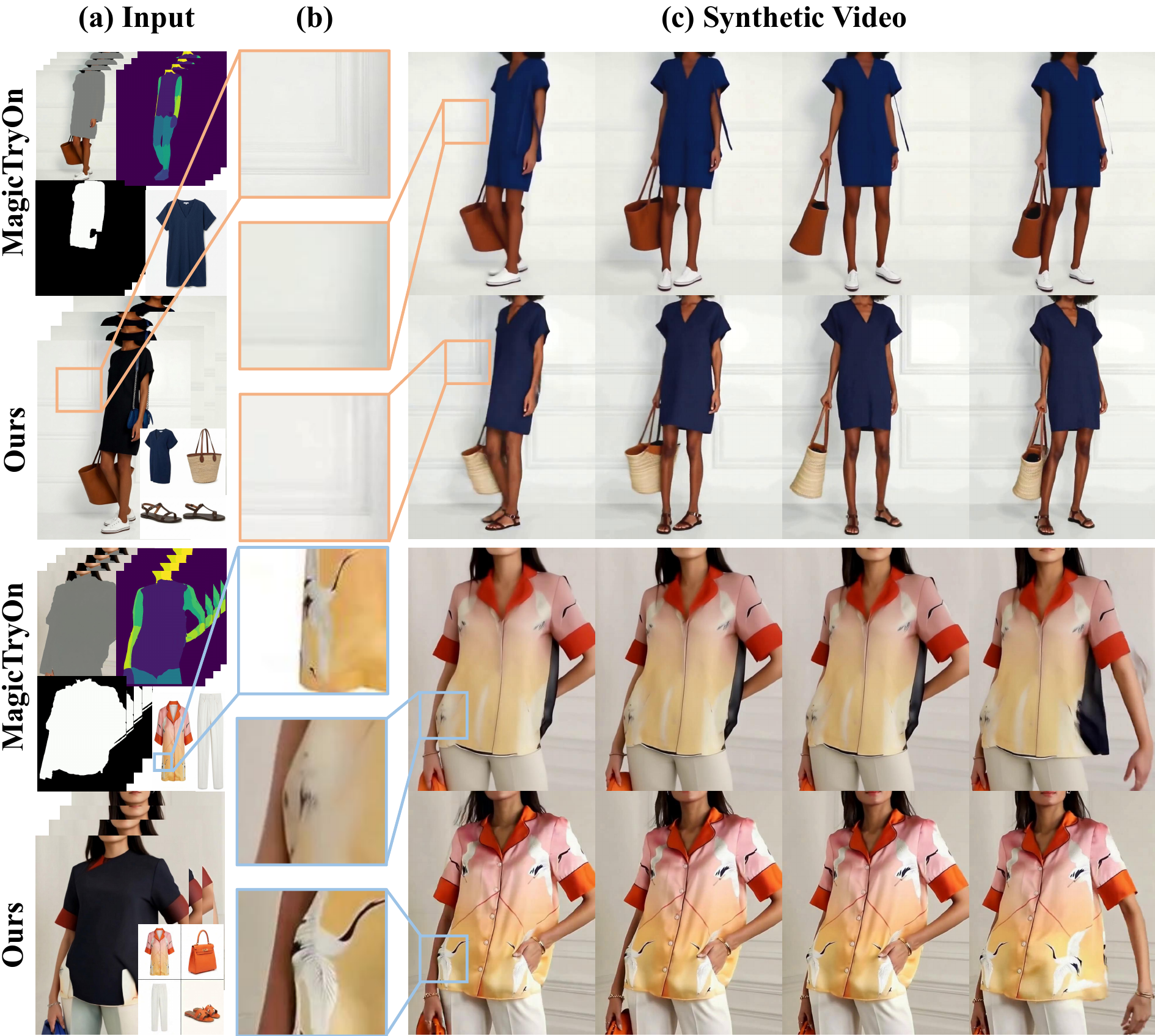}
  \caption{Visual comparison of inputs and synthesis quality.}
  \label{intro}
\end{figure}

\begin{figure*}[t]
  \centering
  \includegraphics[width=0.83\textwidth]{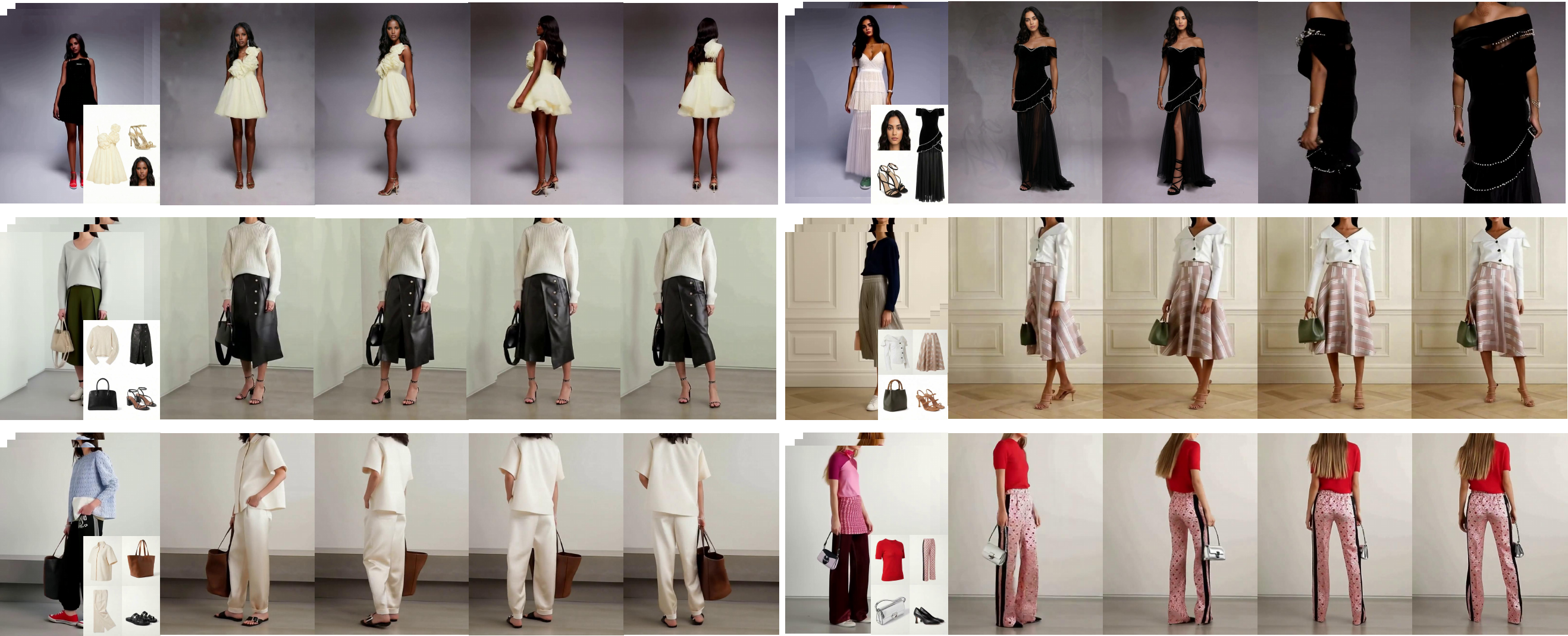}
  \caption{OmniTryOn achieves Try-On Anything by transferring diverse wearable objects in a single inference pass.}
  \label{teaser}
\end{figure*}

Video Virtual Try-On (VVT) has received significant attention due to its extensive applications in e-commerce and digital entertainment. Given a source video of a person and a reference image of the target garment, VVT aims to seamlessly transfer the garment onto the individual while preserving the spatiotemporal consistency of the background and motions. Empowered by the remarkable visual fidelity of recent advanced video generation models~\cite{yang2024cogvideox, kong2024hunyuanvideo, wan2025wan}, state-of-the-art VVT approaches~\cite{li2025magictryon, he2025devil, zeng2025eevee, chong2025catv2ton, fang2024vivid} have achieved impressive generation quality and high-fidelity garment preservation.

Despite these advances, a significant gap remains between existing methods and real-world applications, due to two fundamental limitations. First, current approaches are restricted to single-garment transfer and cannot handle the simultaneous try-on of diverse wearable objects such as garments, handbags, and shoes, as illustrated by the MagicTryOn~\cite{li2025magictryon} results in Fig.~\ref{intro} (c). Second, to enforce spatial localization and structural alignment, existing frameworks heavily rely on explicit geometric priors (e.g., garment-agnostic masks and dense human poses) extracted by a complex pipeline of auxiliary models~\cite{li2020self, wu2019detectron2, AniLines}. Such rigid representations misalign with real physical dynamics and cause artifacts including background distortion and texture collapse, as shown in Fig.~\ref{intro} (b). To bridge these gaps, we introduce the \textbf{Try-On Anything} task, which transfers diverse wearable objects onto a person in a single inference pass, and develop a unified framework free of explicit geometric priors to accomplish it.

To this end, we first introduce \textbf{TryAny-Bench}, a comprehensive benchmark tailored for the Try-On Anything task. TryAny-Bench provides high-fidelity paired videos that extend beyond conventional try-on categories (i.e., garments only),simultaneously  accommodating customization targets such as garments, handbags, shoes, and facial identities. Due to the absence of paired videos, existing VVT datasets~\cite{zeng2025eevee, fang2024vivid, dong2019fw} construct training pairs from destructively masked videos as reference inputs (Fig.~\ref{intro} (a)), which erases essential physical contexts. In contrast, TryAny-Bench provides paired reference and target videos with distinct wearable objects while preserving human motions and background dynamics, removing the need for artificial masking. Alongside the dataset, TryAny-Bench includes a specialized evaluation protocol that comprehensively assesses multidimensional try-on quality, establishing a robust foundation for Try-On Anything.

To comprehensively tackle this task, we present \textbf{OmniTryOn}, a framework free of explicit geometric priors. OmniTryOn addresses two coupled requirements: simultaneously injecting diverse wearable objects and enforcing strict spatiotemporal consistency. For the first, we introduce the \textit{First Frame Wearable Cache}. Inspired by the in-context capabilities of DiTs~\cite{huang2024context, chen2025first, bian2025video}, this strategy formulates the initial frame of the target video as a unified wearable object cache, fine-grained object details continuously propagate to subsequent frames via full attention~\cite{bian2025video}. This consolidated design also scales naturally to a variable number of wearable objects. For the second, we propose the \textit{Spatiotemporally Consistent Rotary Position Embedding} (STC-RoPE), which assigns identical 3D positional coordinates~\cite{su2024roformer} to reference and target tokens, enforcing motion and background consistency. Finally, to mitigate the optimization difficulty of the Try-On Anything task, we employ the \textit{Gradual Try-On} (GTO) training strategy, which progressively extends the model's synthesis capability from conventional garments to diverse wearable objects.
The contributions of our work are:

\begin{itemize}
    \item We propose the novel Try-On Anything task and introduce \textbf{TryAny-Bench}, a benchmark tailored for this paradigm, containing paired videos across diverse wearable objects and a specialized evaluation protocol.
    \item We present \textbf{OmniTryOn}, which eliminates explicit geometric priors via a First Frame Wearable Cache and STC-RoPE. Combined with the GTO training strategy, it transfers diverse wearables in a single inference pass while preserving motion and background consistency.
    \item Extensive experiments show that OmniTryOn consistently outperforms specialized VVT models and general video editing baselines on multiple metrics, with superior visual fidelity and spatiotemporal stability.
\end{itemize}

\begin{figure*}[t]
  \centering
  \includegraphics[width=0.82\linewidth]{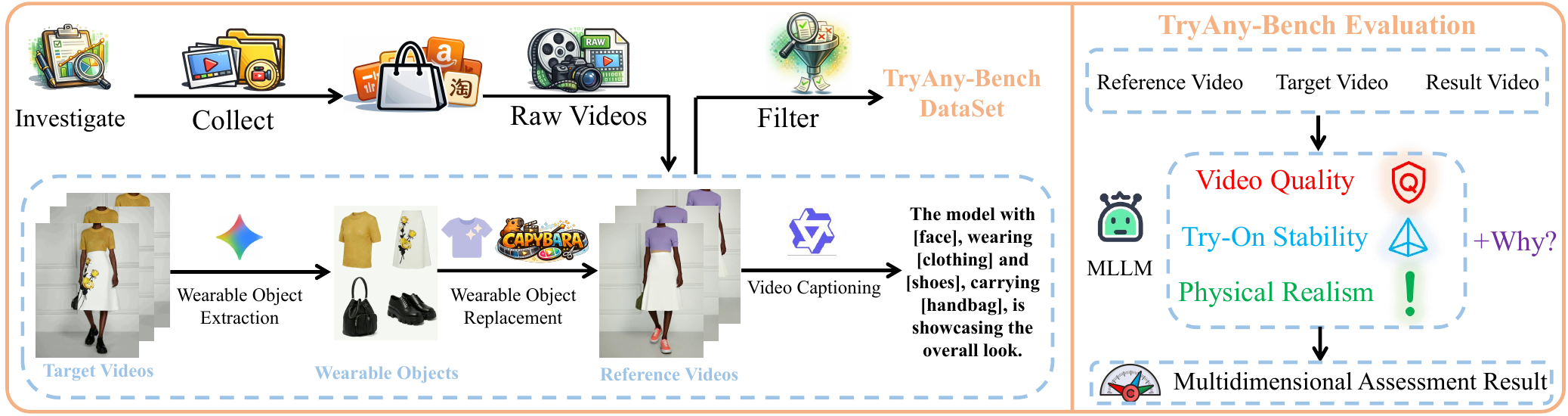}
  \caption{Overview of the TryAny-Bench data construction pipeline and evaluation protocol.}
  \label{data}
\end{figure*}

\section{Related Work}
\subsection{Video Generation}

Diffusion-based video generation has advanced from text-to-video synthesis~\cite{blattmann2023align, yang2024cogvideox} to controllable generation~\cite{he2025fulldit2, lei2025animateanything} and video editing~\cite{yeunified, yang2025unified}. Early works adopt U-Net~\cite{ronneberger2015u} backbones, while recent works build on Diffusion Transformers (DiTs)~\cite{peebles2023scalable, kong2024hunyuanvideo, wan2025wan} to achieve long, high-resolution synthesis. To accommodate diverse control signals such as depth, pose, and segmentation, VACE~\cite{jiang2025vace} unifies multiple conditional inputs into a single generation pipeline. In parallel, Video-As-Prompt~\cite{bian2025video} exploits the in-context capabilities of DiTs~\cite{huang2024context, chen2025first} for semantically controlled generation. Despite this progress, general video generation models remain limited when required to place multiple wearable objects onto specific regions of a dynamic person while simultaneously preserving motion and background dynamics. Building on this line of work, we extend a DiT-based generative framework to explicitly address the Try-On Anything task.

\subsection{Video Virtual Try-On}
Inspired by the success of diffusion-based video generation~\cite{yang2024cogvideox, wan2025wan, kong2024hunyuanvideo}, recent VVT methods~\cite{chong2025catv2ton, he2025devil, zeng2025eevee, zuo2025dreamvvt, fang2024vivid, li2025magictryon} have shifted from U-Net~\cite{ronneberger2015u} (e.g., ViViD~\cite{fang2024vivid}, RealVVT~\cite{li2025realvvt}, Tunnel~\cite{xu2024tunnel}) to DiT~\cite{peebles2023scalable} backbones for enhanced visual realism and detail preservation. These approaches typically follow a multi-branch design, in which one branch encodes garment features while another encodes explicit geometric priors (e.g., garment-agnostic masks and human poses) extracted from auxiliary modules, and the DiT fuses these streams to drive garment transfer. Other concurrent works~\cite{he2025devil} incorporate Multimodal Large Language Models to enhance garment details. However, this reliance on external modules incurs high computational cost and complicates inference; moreover, existing methods are restricted to single-garment transfer per pass and cannot handle non-garment wearables (e.g., handbags, shoes). To address these issues, we introduce OmniTryOn, a framework free of explicit geometric priors that transfers diverse wearables in a single inference pass, without auxiliary extractors or cascaded inference.

\section{Method}
We first introduce TryAny-Bench, a benchmark comprising a paired video dataset and a specialized evaluation protocol for the Try-On Anything task. We then present OmniTryOn, our generative framework free of explicit geometric priors, which combines the First Frame Wearable Cache for injecting fine-grained object details with the Spatiotemporally Consistent Rotary Position Embedding (STC-RoPE) for enforcing consistency of human motion and background. Finally, we describe the Gradual Try-On (GTO) training strategy, which progressively extends the model's synthesis capability from conventional garments to diverse wearable objects.

\subsection{TryAny-Bench: Benchmarking Try-On Anything}
\label{sec:benchmark}
Existing VVT datasets~\cite{dong2019fw, fang2024vivid, zeng2025eevee, he2025devil} lack support for multi-object try-on and do not provide paired reference and target videos. To close this gap, we introduce \textbf{TryAny-Bench}, a benchmark tailored to the Try-On Anything task. As illustrated in Fig.~\ref{data}, TryAny-Bench consists of a paired video dataset and a specialized VQA-based evaluation protocol.

\begin{table*}[t]
  \centering
  \small
  \begin{tabular}{lccccc}
    \toprule
    \textbf{Dataset / Benchmark} & \textbf{Paired Videos} & \textbf{Try-On Scope} & \textbf{Mask-Free} & \textbf{Multi-dim Eval.} & \textbf{No Start Overexposure} \\
    \midrule
    VVT~\cite{dong2019fw} & $\times$ & Garments & $\times$ & $\times$ & $\checkmark$ \\
    ViViD~\cite{fang2024vivid} & $\times$ & Garments & $\times$ & $\times$ & $\times$ \\
    EEVEE~\cite{zeng2025eevee} & $\times$ & Garments & $\times$ & $\times$ & $\checkmark$ \\
    ViT-HD~\cite{he2025devil} & $\times$ & Garments & $\times$ & $\times$ & $\checkmark$ \\
    \midrule
    \textbf{TryAny-Bench} & \checkmark & Diverse Objects & \checkmark & \checkmark & \checkmark \\
    \bottomrule
  \end{tabular}
  \caption{Comparison between TryAny-Bench and existing VVT datasets.}
  \label{tab:dataset_comparison}
\end{table*}

\subsubsection{TryAny-Bench Dataset Construction}
To construct the core dataset of TryAny-Bench, we systematically process raw e-commerce videos to generate high-fidelity paired video samples. Each sample comprises a reference video, a target video, and corresponding diverse wearable object images.

\noindent\textbf{Wearable Object Extraction.} We first curate over 1,500 raw e-commerce videos as target videos. For each raw video, we sample two anchor frames from the middle segment where the person's full body is visible, and use a commercial image model, Gemini 3.1 Flash Image~\cite{gemini31flashimage2026}, to extract the wearable objects and composite them onto a blank background. The extraction prompt intentionally perturbs the scale and orientation of each object, preventing spatial leakage and encouraging the generative model to learn robust localization rather than copy-pasting. Detailed extraction prompts are in the \textit{\textbf{supplementary material}}.

\noindent\textbf{Wearable Object Replacement.} To construct the reference video, we alter the wearables in the target video while preserving the background and motions. We first apply MagicTryOn~\cite{li2025magictryon} to replace conventional garments (e.g., tops and bottoms). Since this geometric-prior-based model leaves residual artifacts and cannot handle non-garment objects, we sequentially feed the intermediate video into a video editing model, CAPYBARA~\cite{capybara2026rao}, which modifies non-garment objects (e.g., handbags and shoes) and inpaints the artifacts left by MagicTryOn, restoring spatiotemporal consistency. For videos with visible faces, we further apply a face-swapping model~\cite{facefusion2023} to alter facial identities. This yields a reference video with entirely different wearables from the target, removing the destructive masking used in existing datasets.

\noindent\textbf{Fine-Grained Video Captioning.} We employ Qwen3-VL-Instruct-8B~\cite{qwen3technicalreport} to caption the target videos. The prompt focuses on the customization targets, producing sentences of the form: "\textit{The model with [description of face], wearing [description of clothing] and [description of shoes], carrying [description of handbag], is showcasing the overall look.}" Prompt details are in the \textit{\textbf{supplementary material}}.

\noindent\textbf{Quality Control and Statistics.} We crop overexposed initial frames and manually filter samples with accumulated editing errors. The final dataset contains 1,460 paired samples, split into 1,243 training and 217 testing videos. Table~\ref{tab:dataset_comparison} compares TryAny-Bench with existing datasets; a detailed pipeline analysis is in the \textit{\textbf{supplementary material}}.

\subsubsection{TryAny-Bench Evaluation Protocol}
While established metrics~\cite{wang2004image, zhang2018unreasonable, unterthiner2018towards} provide foundational video editing assessments, they often struggle to capture the nuanced, multidimensional requirements of complex virtual try-on scenarios. To complement these metrics and achieve a more comprehensive evaluation, TryAny-Bench introduces a specialized Vision Question Answering (VQA)-based evaluation system tailored for the Try-On Anything task. Although recent VQA approaches~\cite{pu2025mvqa} offer fine-grained assessments for general video editing, their criteria are fundamentally not tailored to the unique physical and semantic demands of novel Try-On Anything task. Therefore, we construct a dedicated evaluation system encompassing three primary aspects: 
(1) Video Quality, evaluated through the sub-dimensions of Visual Fidelity (VF), Action Synchronization (AS), and Environment Stability (ES); 
(2) Try-On Stability, comprising Object Integrity (OI), Material Fidelity (MF), and Scale Naturalness (SN); 
and (3) Physical Realism, which includes Temporal Stability (TS), Anatomical Integrity (AI), and Dynamic Plausibility (DP). Across all dimensions, higher scores denote superior synthesis quality. Crucially, the protocol pairs these quantitative scores with textual rationales to maximize interpretability. The detailed evaluation protocol and specific VQA prompts are in the \textit{\textbf{supplementary material}}.

\begin{figure*}[t]
  \centering
  \includegraphics[width=0.82\linewidth]{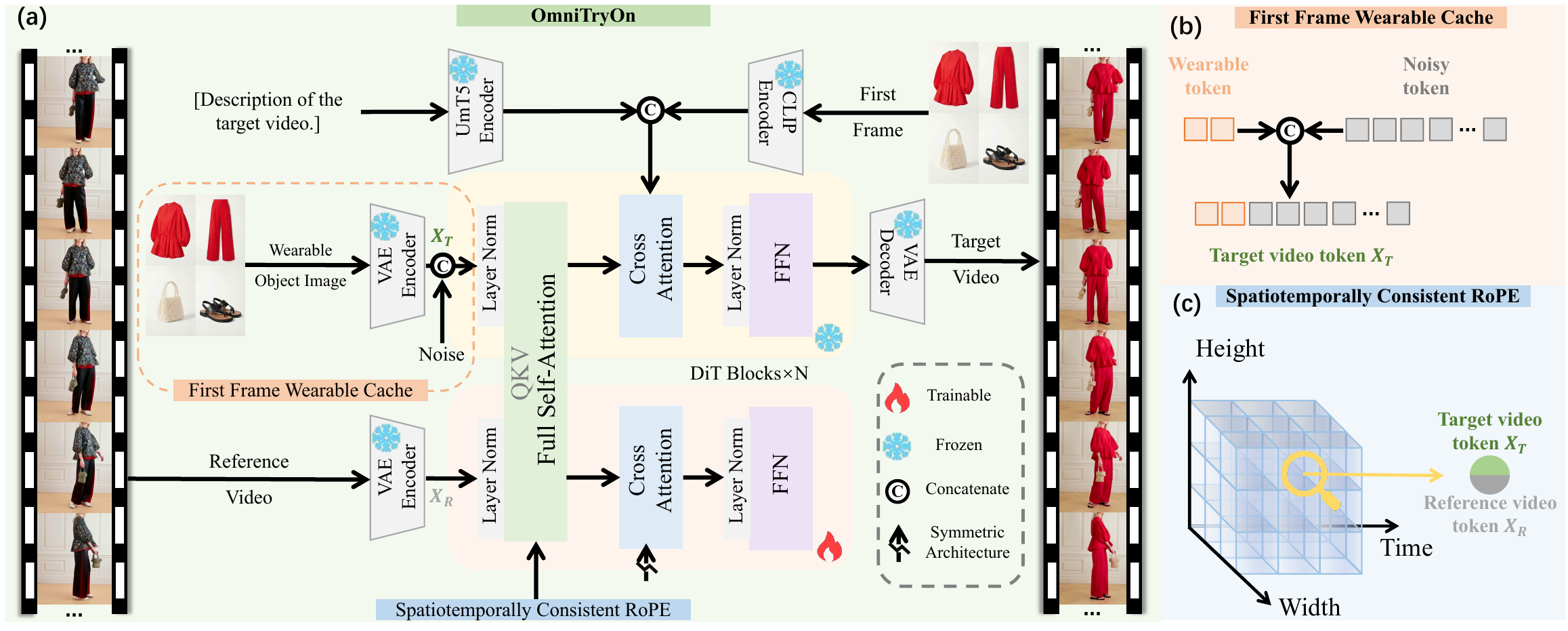}
  \caption{Overview of the OmniTryOn framework.}
  \label{method}
\end{figure*}

\subsection{OmniTryOn}
\label{sec:model}
As illustrated in Fig.~\ref{method} (a), our OmniTryOn framework is built upon Diffusion Transformers (DiTs)~\cite{peebles2023scalable} to accomplish the Try-On Anything task in a single inference pass, free of explicit geometric priors. The framework takes as input a reference video, a wearable object image, and a corresponding text description. The reference video is encoded into the latent space by a VAE encoder to preserve human motion and background dynamics, while the wearable object image is encoded and prepended to the noisy latent sequence as the first frame. In parallel, the text description and the wearable object image are encoded into text and CLIP tokens to provide semantic conditioning. After multiple denoising steps in the DiT backbone, the VAE decoder reconstructs the target video from the resulting latent.

\subsubsection{First Frame Wearable Cache}
As illustrated in Fig.~\ref{method} (b), we prepend the latent of the wearable object image to the noisy target video latents as the first frame, and duplicate the first frame of the reference video once to match the temporal length of the target. Both videos are then flattened into patch tokens $X_T \in \mathbb{R}^{L_{Target} \times d}$ and $X_R \in \mathbb{R}^{L_{Ref} \times d}$ and processed via full attention~\cite{bian2025video} within the DiT blocks:
\begin{equation}
\text{Attention}(Q, K, V) = \text{softmax}\left(\frac{QK^T}{\sqrt{d}}\right)V
\label{eq:attention}
\end{equation}
where $Q = [W_Q^{Target}X_T; W_Q^{Ref}X_R]$, $K = [W_K^{Target}X_T; W_K^{Ref}X_R]$, and $V = [W_V^{Target}X_T; W_V^{Ref}X_R]$ represent the concatenated Queries, Keys, and Values derived from $X_T$ and $X_R$. Benefiting from the robust in-context modeling capabilities of DiTs~\cite{huang2024context, chen2025first, bian2025video}, the prepended first frame of the target video seamlessly functions as a wearable object cache during this attention computation. This cache continuously propagates fine-grained texture and structure of the wearable objects to the subsequent frames, while the reference video tokens supply the conditioning that enforces consistency of complex human motions and background dynamics.

\subsubsection{Spatiotemporally Consistent RoPE}
Preserving inter-frame consistency of the wearable objects while retaining complex human motions and background dynamics, free of explicit geometric priors, is highly challenging. To address this, we propose the Spatiotemporally Consistent Rotary Position Embedding (STC-RoPE), which grounds the reference video, the wearable object image, and the target video in a shared spatiotemporal coordinate system.

DiT-based generative models typically apply 3D RoPE $(t, x, y)$ to video tokens~\cite{su2024roformer}, and these coordinates govern the attention interactions between reference and target tokens. To prevent spatiotemporal overlap between the two streams, prior in-context designs introduce a temporal~\cite{bian2025video} or spatial~\cite{tan2025ominicontrol, wu2025less} offset to the coordinates of reference tokens. This design suits tasks such as style transfer, where the reference only supplies a semantic anchor. In Try-On Anything, however, the target video is required to reproduce the reference motion and background token by token, and to differ only in the designated wearable objects.

STC-RoPE therefore assigns completely identical 3D RoPE $(t, x, y)$ to the reference and target videos, where $t \in [0, f-1]$, $x \in [0, h-1]$, and $y \in [0, w-1]$. Here, $f$, $h$, and $w$ denote the temporal length, height, and width of the latent representations, respectively. Grounding all inputs in this shared spatiotemporal coordinate system establishes a strict identity mapping between reference and target for human motions and backgrounds, and provides consistent spatiotemporal anchors along which the model localizes and transfers the fine-grained features of the wearable objects.

\subsection{Gradual Try-On Training Strategy}
\label{sec:training}
We employ the proposed Gradual Try-On (GTO) training strategy to optimize our OmniTryOn, steadily extending the model's try-on capability from conventional garments to diverse wearables to accomplish the Try-On Anything task. The GTO strategy is decoupled into two progressive stages.

Stage 1. This stage focuses exclusively on conventional garment transfer. During this stage, each paired reference and target video contains different garments but retains identical non-garment wearables. These training pairs are derived from the intermediate outputs of the TryAny-Bench construction pipeline. Compared with other wearable objects, garments occupy larger regions and exhibit highly complex non-rigid deformations, such as dynamic wrinkles and intricate human-garment interactions, making them the most challenging to synthesize. Stage 1 isolates this difficulty, focusing the model's capacity on robust garment transfer.

Stage 2. This stage expands the training scope to include the simultaneous try-on of diverse wearable objects, such as handbags and shoes. Building on the garment transfer capability acquired in Stage 1, the model efficiently learns to localize and generate these semi-rigid objects, while further reinforcing its garment synthesis quality. Through this progressive schedule, the framework ultimately achieves robust multi-object try-on in a single inference pass.

Across both stages, the model is optimized with a flow matching objective~\cite{lipman2022flow}. Let $X_1$ be the target video latent, $X_0 \sim \mathcal{N}(0, \mathbf{I})$ the Gaussian noise, and $X_t = t X_1 + (1 - t) X_0$ the noisy latent at timestep $t \in [0, 1]$. The model $v_\theta$ is trained to predict the velocity field $V_t = X_1 - X_0$:
\begin{equation}
\mathcal{L} = \mathbb{E}_{t, X_0, X_1} \left[ \left\| V_t - v_\theta(X_t, t, C_{Ref}, C_{Text}) \right\|_2^2 \right]
\label{eq:loss}
\end{equation}
where $C_{Ref}$ and $C_{Text}$ denote the embeddings extracted from the reference video and the text captions, respectively.

\section{Experiments}
\subsection{Experimental Settings}
\subsubsection{Implementation Details}
\label{sec:implementation}
We adopt the pretrained weights from Video-As-Prompt~\cite{bian2025video} as the foundational model. All training processes are exclusively conducted on the training set of our curated TryAny-Bench dataset. Following the proposed GTO training strategy, the model is trained for two epochs, dedicating one epoch to each stage. During training, each video sequence contains 49 frames, and the batch size is set to 1 per GPU. The parameters are optimized using the AdamW~\cite{loshchilov2017decoupled} optimizer with a learning rate of $1 \times 10^{-5}$. All experiments are executed on 8 NVIDIA H200 GPUs. For efficiency and robustness, we fine-tune only one quarter of the DiT blocks; detailed configurations are in the \textit{\textbf{supplementary material}}.

\subsubsection{Evaluation Metrics}
\label{sec:metrics}
To comprehensively assess the performance of our framework, we employ a dual evaluation system comprising established quantitative metrics and the multidimensional protocol introduced in TryAny-Bench.

\subsubsection{Established Metrics.} Following previous VVT works~\cite{li2025magictryon, he2025devil, chong2025catv2ton, fang2024vivid}, we adopt four widely used metrics: SSIM~\cite{wang2004image}, LPIPS~\cite{zhang2018unreasonable}, $\text{VFID}_I$~\cite{fang2024vivid}, and $\text{VFID}_R$~\cite{fang2024vivid}. Specifically, SSIM measures the structural similarity between the generated and reference videos, while LPIPS captures fine-grained perceptual differences. Furthermore, $\text{VFID}_I$ (using the I3D~\cite{carreira2017quo} backbone) and $\text{VFID}_R$ (using the ResNeXt~\cite{xie2017aggregated} backbone) are utilized to simultaneously evaluate the spatial visual quality and temporal consistency of the synthesized videos.

\noindent\textbf{VQA-Based Evaluation.} To execute the specialized VQA-based evaluation in our experiments, we employ the advanced \textit{gemini-3-flash-preview}~\cite{gemini3flashpreview} as our evaluator. To ensure deterministic and reproducible scoring, the sampling temperature of the evaluator is set to 0.0. 

\subsection{Quantitative Comparison}
We evaluate OmniTryOn against three specialized VVT baselines (ViViD~\cite{fang2024vivid}, CatV$^2$TON~\cite{chong2025catv2ton}, and MagicTryOn~\cite{li2025magictryon}) and two general video editing models (VACE~\cite{jiang2025vace} and VAP~\cite{bian2025video}) on the TryAny-Bench test set. As reported in Table~\ref{tab:comparison_results}, OmniTryOn outperforms all baselines across every established metric, confirming its capability to simultaneously transfer diverse wearable objects while preserving human motion and background consistency.

\begin{table}[t]
  \centering
  \fontsize{8.3pt}{9pt}\selectfont
  \setlength{\tabcolsep}{0.5pt}
  \begin{tabular}{lccccc}
    \toprule
    Method & VFID$_I\downarrow$ & VFID$_R\downarrow$ & SSIM$\uparrow$ & LPIPS$\downarrow$ & \makecell{Inference \\ Time (s)$\downarrow$} \\
    \midrule
    Magic-TryOn\cite{li2025magictryon} & 18.256 & 0.654 & \underline{0.820} & \underline{0.162} & 356 \\
    CatV2TON\cite{chong2025catv2ton}  & 24.141 & 2.172 & 0.793 & 0.201 & 220 \\
    ViViD\cite{fang2024vivid}      & 19.139 & \underline{0.454} & 0.787 & 0.190 & 216 \\
    VACE\cite{jiang2025vace}       & \underline{17.975} & 1.653 & 0.801 & 0.173 & 217 \\
    VAP\cite{bian2025video}        & 41.767 & 5.763 & 0.651 & 0.419 & \underline{186} \\
    \midrule
    \textbf{Ours} & \textbf{14.704} & \textbf{0.235} & \textbf{0.821} & \textbf{0.144} & \textbf{186} \\
    \bottomrule
  \end{tabular}
  \caption{Quantitative comparison results. Inference time is the average on the TryAny-Bench test set; for baselines, includes overhead of prior extraction and cascaded inference.}
  \label{tab:comparison_results}
\end{table}

We attribute the gap between OmniTryOn and the specialized VVT baselines to three intrinsic limitations of the baselines. First, their garment-agnostic input discards the physical detail needed for authentic appearance synthesis. Second, multi-object transfer requires cascaded inference across garments, along which spatiotemporal errors accumulate. Third, they are architecturally restricted to garments and cannot synthesize non-garment wearables. OmniTryOn resolves all three in a single inference pass: the First Frame Wearable Cache injects the conditions for diverse wearable objects at once, and STC-RoPE preserves complex human motions and background dynamics. Against the general video editing baselines, OmniTryOn also leads on every metric; in particular, VAP~\cite{bian2025video} yields the poorest overall performance, indicating that general video editing models cannot maintain the strict spatiotemporal consistency required by Try-On Anything.

\begin{figure}[h]
  \centering
  \includegraphics[width=0.83\linewidth]{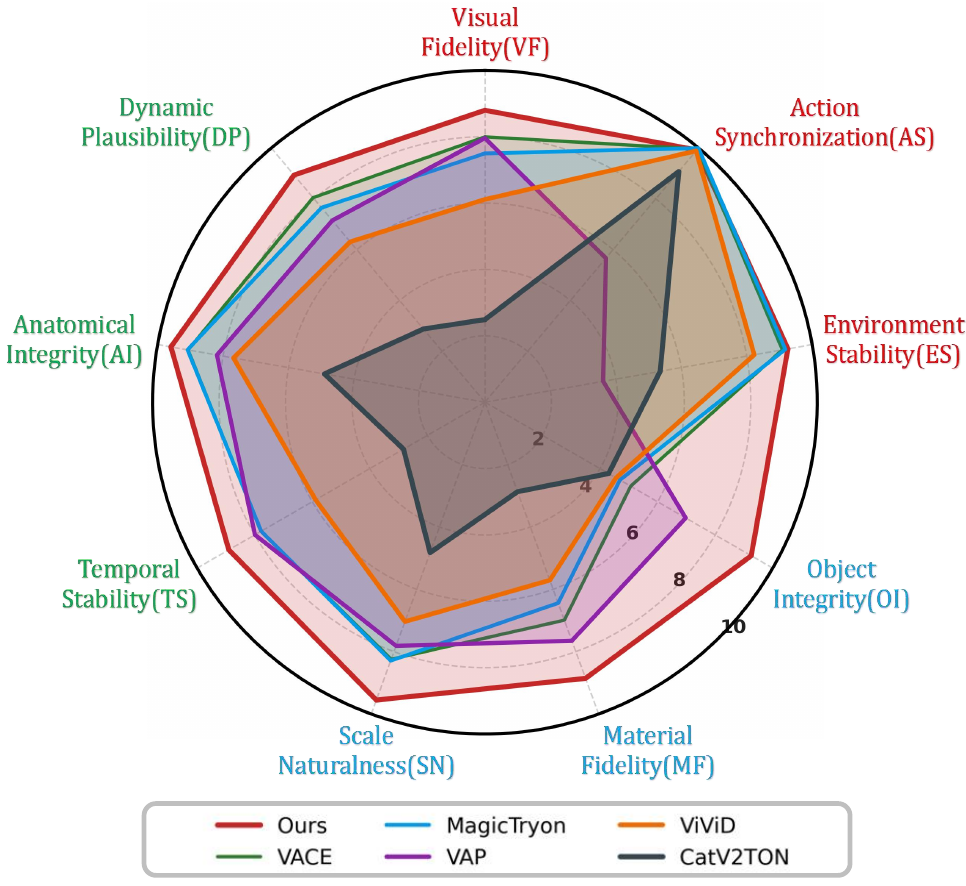}
  \caption{Comprehensive VQA-based evaluation results.}
  \label{fig:radar_chart}
\end{figure}

To further validate the multidimensional superiority of our model beyond established metrics, we conduct a VQA-based evaluation; fine-grained results are shown in Fig.~\ref{fig:radar_chart}. OmniTryOn consistently outperforms all baselines across every dimension, with largest gains on challenging and critical criteria (Object Integrity and Material Fidelity). Numerical values are in the \textit{\textbf{supplementary material}}. To validate the reliability of this VQA protocol, we further conduct a user study; humans consistently prefer OmniTryOn over all baseline, and the ranking matches the VQA ranking. The full results are in the \textit{\textbf{supplementary material}}.

\begin{figure*}[t]
  \centering
  \includegraphics[width=0.85\linewidth]{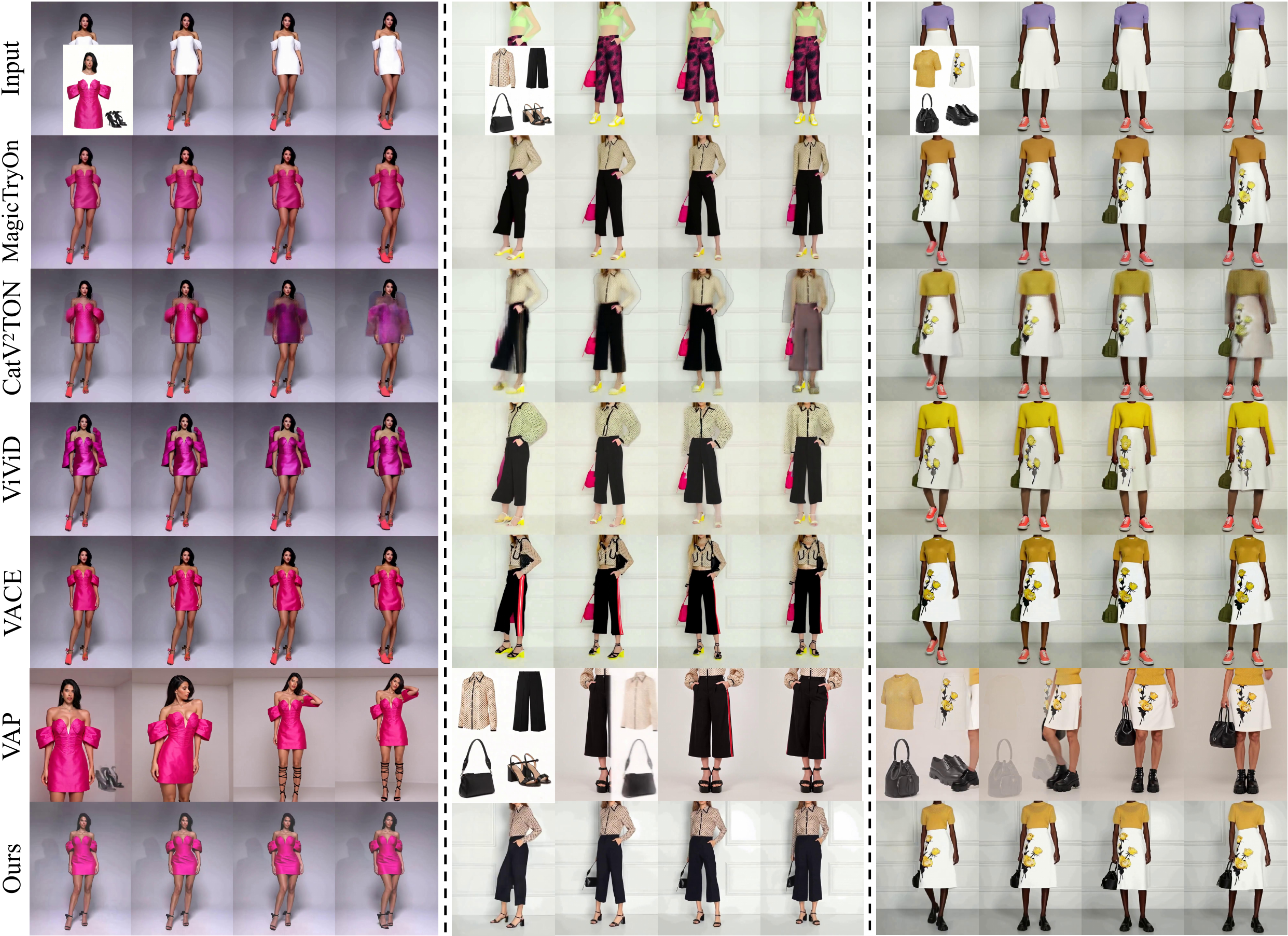}
  \caption{Qualitative comparison between OmniTryOn and state-of-the-art baselines.}
  \label{fig:qualitative}
\end{figure*}

\subsection{Qualitative Comparison}
Figure~\ref{fig:qualitative} presents a visual comparison between OmniTryOn and the state-of-the-art baselines. Existing VVT models fail to synthesize non-garment wearable objects, and even for conventional garments they suffer from severe visual degradations: CatV$^2$TON~\cite{chong2025catv2ton} produces ghosting artifacts and boundary residues from destructive garment-agnostic masking, while ViViD~\cite{fang2024vivid} alters the target attributes with unintended style modifications and color deviations. General video editing models suffer from a different failure mode: VACE~\cite{jiang2025vace} hallucinates spurious visual features such as arbitrary stripes on unrelated garments, and VAP~\cite{bian2025video} undergoes a catastrophic spatiotemporal collapse, degrading into fragmented, floating layouts that no longer preserve the original human motion or background.

In contrast, OmniTryOn transfers diverse wearable objects with high fidelity, preserving the high-frequency structural details of the target objects while maintaining robust spatiotemporal consistency for both human motions and backgrounds. We further validate OmniTryOn on long 96-frame sequences with fast motion and heavy occlusion, and on in-the-wild scenes drawn from outside the e-commerce domain; both settings are visualized in the \textit{\textbf{supplementary material}} and demonstrate the robustness of the First Frame Wearable Cache to temporal length and domain shift.

\subsection{Ablation Study}
\subsubsection{Cross-Benchmark Comparison}
Since no existing dataset supports the Try-On Anything task, our main experiments are conducted on TryAny-Bench. To further evaluate the generalization capabilities of OmniTryOn, we first conduct a garment-only evaluation on intermediate outputs of the TryAny-Bench pipeline. Additionally, we extend our evaluation to two public VVT benchmarks, ViViD~\cite{fang2024vivid} and EEVEE~\cite{zeng2025eevee}, focusing our comparison specifically against the most competitive open-source state-of-the-art specialist models. As reported in Table~\ref{tab:comparison_results_v2}, in the garment-only setting, OmniTryOn is competitive with the top garment-specialist baseline on the discriminative metrics, confirming that OmniTryOn generalizes well to garment-only scenarios and to data unseen during training.

\begin{table}[!t]
  \centering
  \setlength{\tabcolsep}{1pt}
  \fontsize{9pt}{9pt}\selectfont
  \begin{tabular}{lcccc}
    \toprule
    Method & VFID$_I\downarrow$ & VFID$_R\downarrow$ & SSIM$\uparrow$ & LPIPS$\downarrow$ \\
    \midrule
    \multicolumn{5}{l}{\textit{Garment-Only (TryAny-Bench)}} \\
    Magic-TryOn\cite{li2025magictryon} & \underline{12.228} & \underline{0.251} & \textbf{0.858} & \underline{0.118} \\
    CatV2TON\cite{chong2025catv2ton}    & 21.952 & 3.029 & 0.833 & 0.159 \\
    ViViD\cite{fang2024vivid}      & 16.167 & 0.332 & 0.821 & 0.152 \\
    VACE\cite{jiang2025vace}       & 13.613 & 0.604 & 0.846 & 0.128 \\
    VAP\cite{bian2025video}        & 50.804 & 9.390 & 0.629 & 0.435 \\
    \textbf{Ours} & \textbf{12.152} & \textbf{0.170} & \underline{0.851} & \textbf{0.117} \\
    \midrule
    \multicolumn{5}{l}{\textit{ViViD}} \\
    ViViD\cite{fang2024vivid}      & 10.39 & 0.253 & \underline{0.862} & 0.124 \\
    Magic-TryOn\cite{li2025magictryon} & \underline{9.49}  & \underline{0.205} & \underline{0.903} & \underline{0.094} \\
    \textbf{Ours} & \textbf{10.64} & \textbf{0.192} & \textbf{0.905} & \textbf{0.092} \\
    \midrule
    \multicolumn{5}{l}{\textit{EEVEE}} \\
    ViViD\cite{fang2024vivid}      & 14.30 & 0.714 & 0.855 & 0.156 \\
    Magic-TryOn\cite{li2025magictryon} & \textbf{10.82} & \underline{0.710} & \textbf{0.916} & \underline{0.113} \\
    \textbf{Ours} & \underline{13.23} & \textbf{0.439} & \underline{0.878} & \textbf{0.095} \\
    \bottomrule
  \end{tabular}
  \caption{Cross-benchmark comparison. The upper block reports garment-only evaluation, whose test pairs are drawn from intermediate outputs of the TryAny-Bench construction pipeline. The middle and lower blocks report evaluations on 256-sample subsets of the ViViD~\cite{fang2024vivid} and EEVEE~\cite{zeng2025eevee} benchmarks.}
  \label{tab:comparison_results_v2}
\end{table}

\subsubsection{Spatiotemporally Consistent RoPE}
To validate the necessity of STC-RoPE, we compare it against three variants that introduce positional biases between reference and target tokens: Spatial-biased RoPE~\cite{wu2025less}, Temporal-biased RoPE~\cite{bian2025video}, and a combined Spatiotemporal-biased RoPE (detailed configurations in the \textit{\textbf{supplementary material}}). For efficiency, all variants are evaluated after Stage 1 of GTO. As reported in Table~\ref{tab:ablation_rope}, STC-RoPE, which introduces no positional bias, yields the best performance, confirming that assigning identical positional coordinates enforces the strict identity mapping of backgrounds and motions required by Try-On Anything.

\begin{table}[htbp]
  \centering
  \resizebox{\linewidth}{!}{
    \begin{tabular}{lcccc}
      \toprule
      Method & VFID$_I\downarrow$ & VFID$_R\downarrow$ & SSIM$\uparrow$ & LPIPS$\downarrow$ \\
      \midrule
      Temporal-biased RoPE & 12.842 & 0.173 & 0.847 & 0.120 \\
      Spatial-biased RoPE & 21.910 & 1.902 & 0.676 & 0.330 \\
      Spatiotemporal-biased RoPE & 22.891 & 2.872 & 0.672 & 0.326 \\
      \midrule
      \textbf{STC-RoPE} & \textbf{12.152} & \textbf{0.170} & \textbf{0.851} & \textbf{0.117} \\
      \bottomrule
    \end{tabular}
  }
  \caption{Ablation study of different RoPE variants.}
  \label{tab:ablation_rope}
\end{table}

\subsubsection{Gradual Try-On Training Strategy}
To validate the GTO training strategy, we compare it against a variant that jointly optimizes for all wearable objects from scratch (Direct strategy) under the same number of epochs. As reported in Table~\ref{tab:strategy_comparison}, GTO yields superior discriminative metrics, confirming that progressively acquiring the try-on capability leads to a more stable optimization.

\begin{table}[htbp]
  \centering
  \resizebox{\linewidth}{!}{
  \begin{tabular}{lcccc}
    \toprule
    Method & VFID$_I\downarrow$ & VFID$_R\downarrow$ & SSIM$\uparrow$ & LPIPS$\downarrow$ \\
    \midrule
    Direct Strategy & \textbf{14.679} & 0.264 & 0.816 & 0.148 \\
    GTO Strategy    & 14.704 & \textbf{0.235} & \textbf{0.821} & \textbf{0.144} \\
    \bottomrule
  \end{tabular}
  }
  \caption{Ablation study of the training strategy.}
  \label{tab:strategy_comparison}
\end{table}

\section{Conclusion}
We present a comprehensive solution to the Try-On Anything task, comprising the TryAny-Bench benchmark and the OmniTryOn model, which is free of explicit geometric priors. OmniTryOn injects wearables through a First Frame Wearable Cache and preserves motion and background with STC-RoPE, while GTO training strategy progressively extends synthesis from garments to diverse customization targets. Experiments show that OmniTryOn achieves superior visual fidelity and stability, providing a strong baseline for future work on immersive try-on.

\bibliography{aaai2027}

\clearpage

\begin{figure*}[t]
  \centering
  \includegraphics[width=0.7\linewidth]{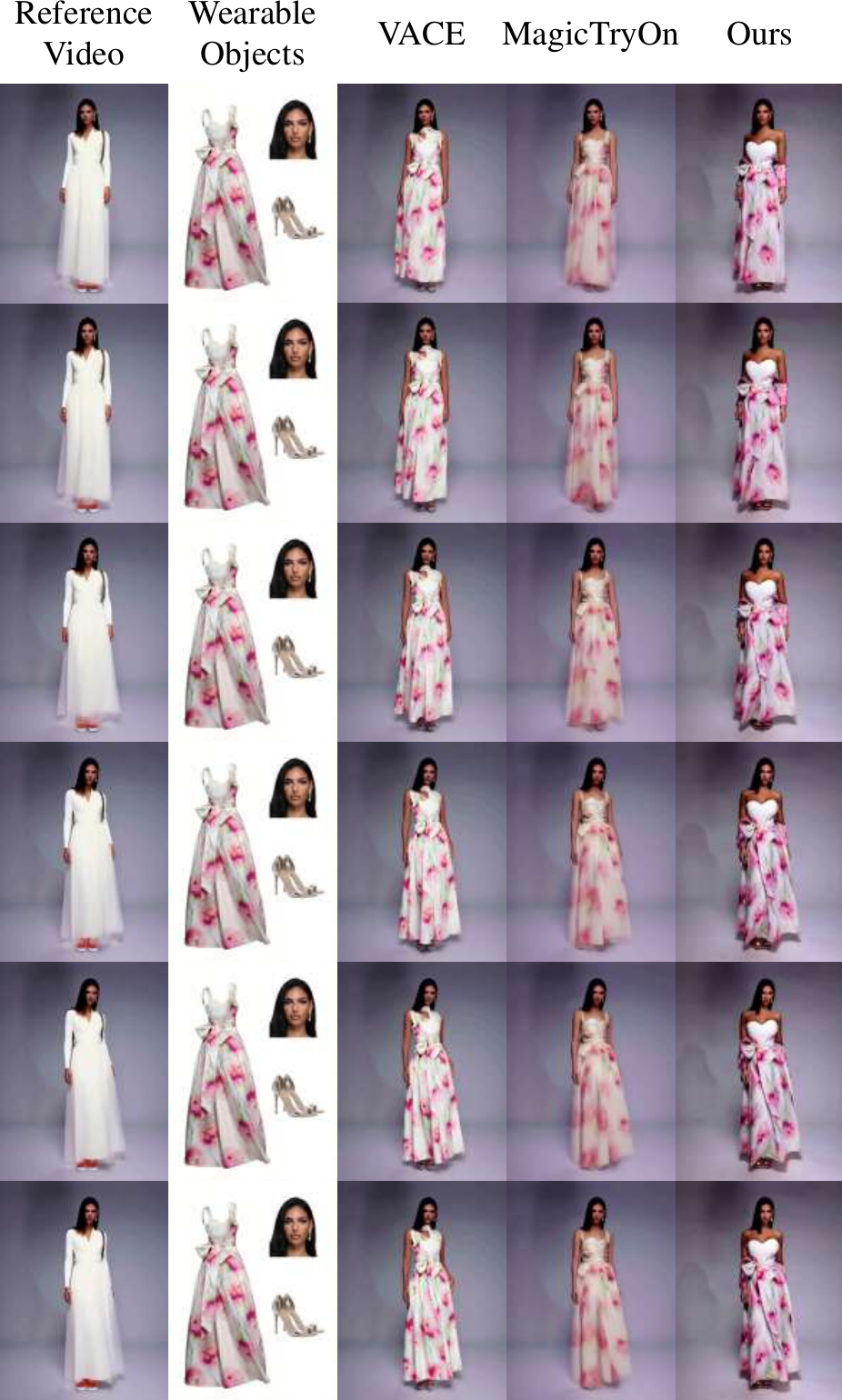}
  \caption{Additional visualization results. Please zoom in for more details.}
  \label{fig1}
\end{figure*}

\begin{figure*}[t]
  \centering
  \includegraphics[width=0.7\linewidth]{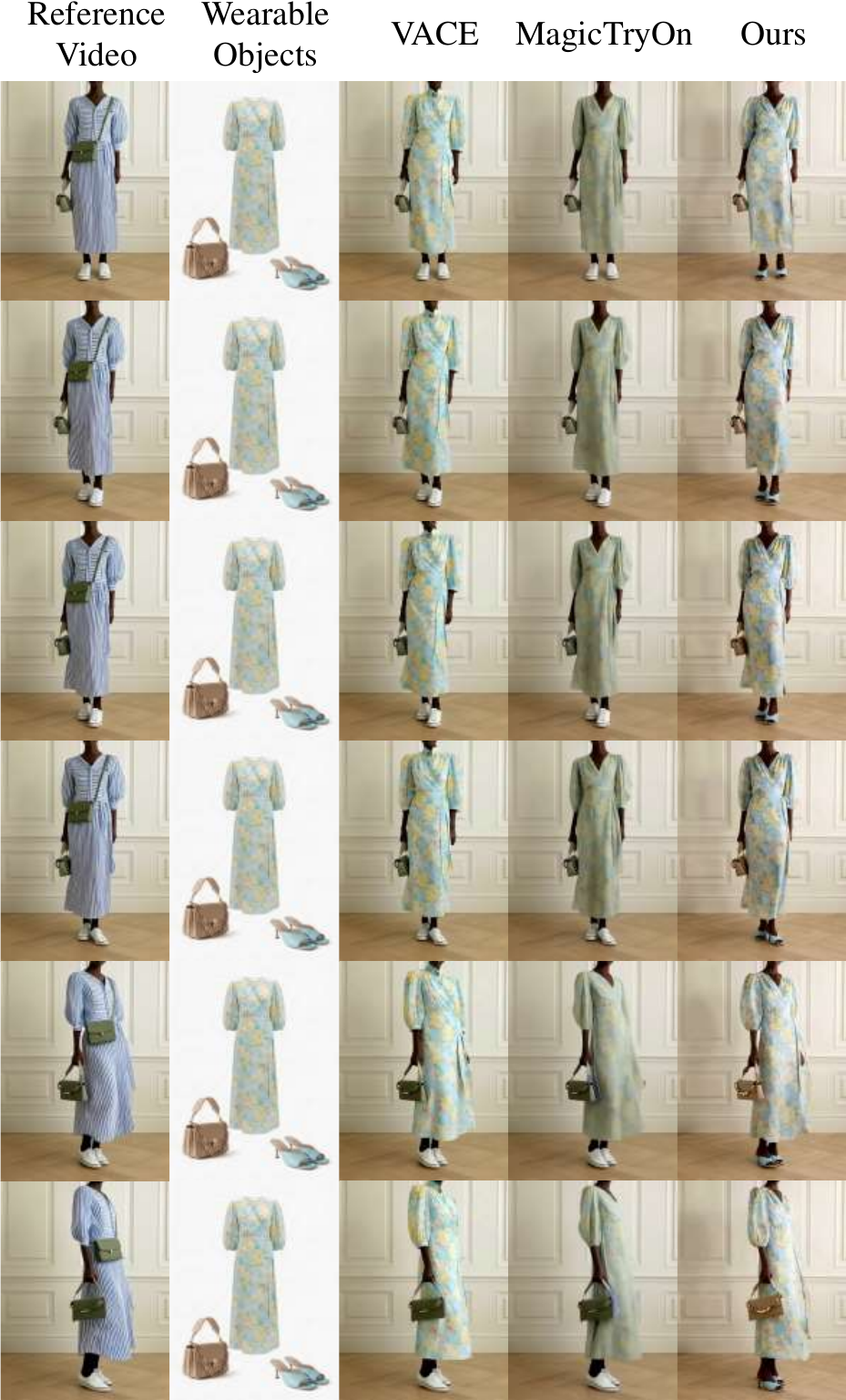}
  \caption{Additional visualization results. Please zoom in for more details.}
  \label{fig2}
\end{figure*}

\begin{figure*}[t]
  \centering
  \includegraphics[width=0.7\linewidth]{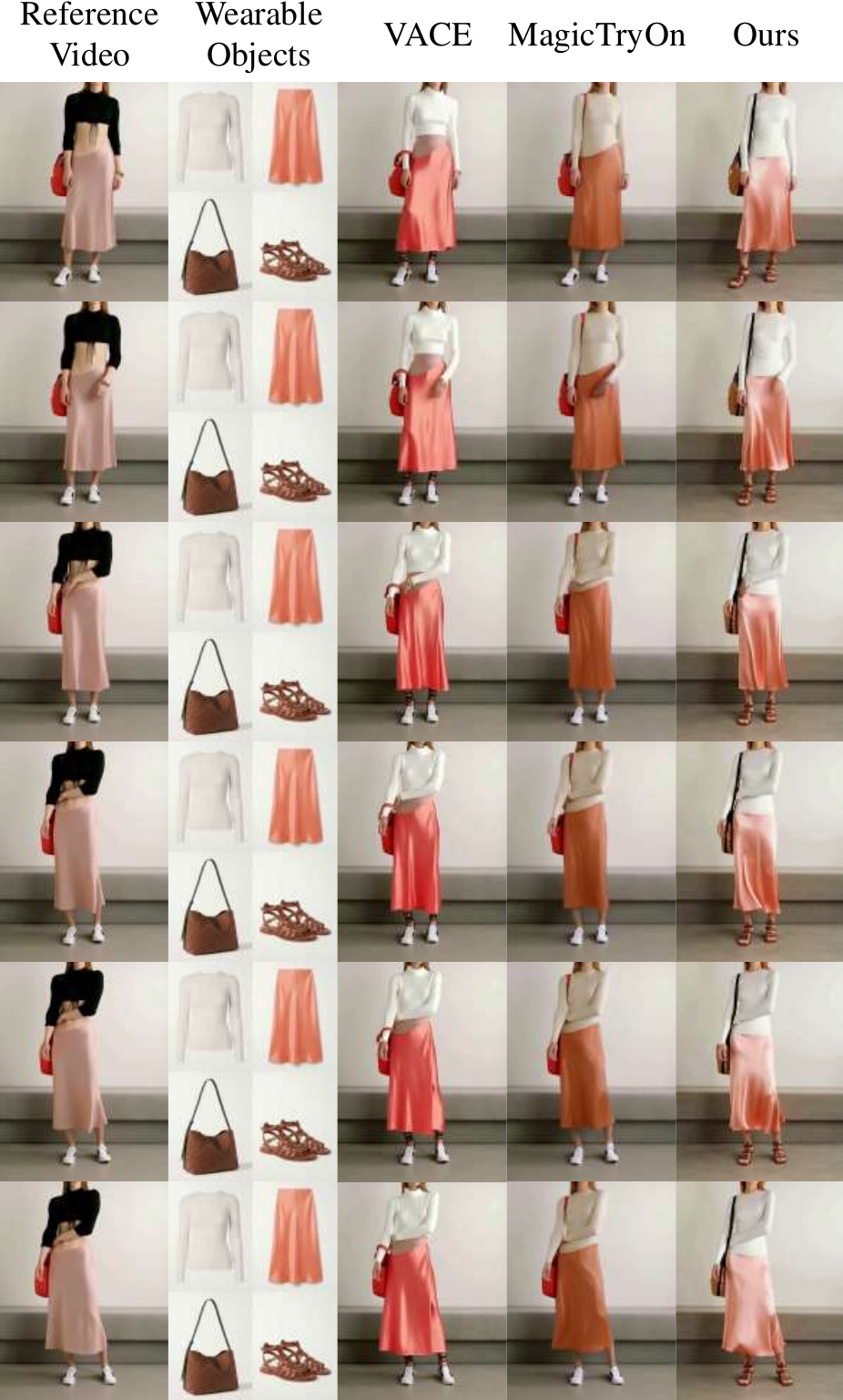}
  \caption{Additional visualization results. Please zoom in for more details.}
  \label{fig3}
\end{figure*}

\clearpage

\pdfinfo{
/TemplateVersion (2027.1)
}

\setcounter{secnumdepth}{0} 

%


\title{OmniTryOn: Video Try-On Anything at Once! \\ \textemdash Supplementary Materials}
\author{
    Written by AAAI Press Staff\textsuperscript{\rm 1}\thanks{With help from the AAAI Publications Committee.}\\
    AAAI Style Contributions by Peter Patel Schneider,
    Sunil Issar,\\
    J. Scott Penberthy,
    George Ferguson,
    Hans Guesgen,
    Francisco Cruz\equalcontrib\corresponding,
    Marc Pujol-Gonzalez\equalcontrib\corresponding
}
\affiliations{
    \textsuperscript{\rm 1}Association for the Advancement of Artificial Intelligence\\


    1101 Pennsylvania Ave, NW Suite 300\\
    Washington, DC 20004 USA\\
    proceedings-questions@aaai.org
%
}

\maketitle


This appendix provides more implementation details, additional details and analysis to complement the main paper. The content is organized as follows:
\begin{itemize}
\item More Implementation Details
\item Human Study for VQA Protocol Reliability
\item Detailed Quantitative Results of TryAny-Bench
\item Analysis of the Dataset Construction Pipeline
\item Additional Visualization Results
\end{itemize}


\section{More Implementation Details}
\label{sec: More Implementation Details}
In this section, we provide extended implementation details essential for reproducing our data curation pipeline and evaluation framework. Specifically, we present the exact prompts formulated to guide the image model in extracting diverse wearable objects, alongside those used by the multimodal large language model (MLLM) to generate fine-grained semantic video captions. These automated data processing steps are pivotal for constructing the high-fidelity TryAny-Bench dataset. Furthermore, we detail the comprehensive Vision Question Answering (VQA) system prompt that drives our multidimensional evaluation protocol, ensuring a rigorous, interpretable, and reproducible assessment of the Try-On Anything task. Finally, we describe the parameter-efficient fine-tuning configuration adopted for OmniTryOn and the explicit coordinate assignments of the RoPE variants used in the ablation study of the main paper.
\subsection{Detailed Prompt for Wearable Object Extraction}
In the data curation pipeline, we utilize Gemini 3.1 Flash Image~\cite{gemini31flashimage2026} to extract wearable objects and the model's facial identity from raw e-commerce videos. To ensure the generative model learns robust spatial adaptation rather than trivial copying, the prompt is carefully designed to instruct the image model to synthesize views, avoid duplicates, and arrange the items in a compact knolling layout. We also specifically enforce high visual fidelity for complex items like footwear. The complete prompt is as follows:
\begin{tcolorbox}[title={Prompt for Wearable Object Extraction}, breakable, pad at break*=1mm]

Deconstruct the fashion model image into a variable number of isolated product shots (e.g., upper/lower garments, dresses, bags, shoes) and extract the model's face as a forward-facing professional beauty headshot. Completely remove the model's body, skin, and limbs from the clothing items. Synthesize any multiple reference views into a single, definitive set of unique items to avoid duplicates.

Arrange all extracted items and the headshot into a compact, non-overlapping knolling layout on a pure white background as high-quality e-commerce product photography. Present the garments in a "volumetric invisible mannequin" or "flat lay" style. Choose the optimal camera perspective for each individual item to best showcase its details, texture, and three-dimensional shape. Strictly maintain the original texture, details, and visual fidelity of all items, with particular emphasis on footwear.
\end{tcolorbox}

\subsection{Detailed Prompt for Fine-Grained Video Captioning}
To provide accurate semantic guidance for the generative framework, we employ Qwen3-VL-Instruct-8B~\cite{qwen3technicalreport} to caption the target videos. The prompt is strictly formatted to ensure the multimodal model focuses on the customization targets (face, clothing, shoes, and handbags) and outputs a consistent sentence structure. The prompt used for caption generation is detailed below:
\begin{tcolorbox}[title={Prompt for Fine-Grained Video Captioning}]
Analyze the provided video of a fashion model. Your task is to generate a fine-grained, highly accurate semantic description of the model's appearance and the specific wearable objects they are showcasing. Pay close attention to the details, textures, colors, and structural styles of the items.

You must strictly output the description using the exact following template: "The model with [description of face], wearing [description of clothing] and [description of shoes], carrying [description of handbag], is showcasing the overall look."

Replace the bracketed placeholders with your detailed observations. Ensure the descriptions are concise yet comprehensive. If a specific category (e.g., handbag) is not present in the video, adjust the template naturally to omit that specific clause while maintaining the overall sentence flow.
\end{tcolorbox}

\subsection{TryAny-Bench Evaluation Protocol and VQA Prompt}
To achieve a comprehensive and interpretable evaluation of the Try-On Anything task, we introduce a specialized VQA-based protocol. This protocol rigorously assesses the generated videos across three primary dimensions: Video Quality, Try-On Stability, and Physical Realism. The MLLM acts as an expert evaluator, providing both a quantitative score (1-10) and a textual rationale for each of the 9 sub-metrics. The complete system prompt used for this evaluation is provided below:

\begin{tcolorbox}[title={Professional VTON Multi-Metric Comprehensive Evaluation Prompt}, breakable, pad at break*=1mm]
\small
\setlength{\parskip}{0.8em} 


\textbf{Expert Identity:}
You are a professional digital artist and computer vision expert. Your role is to perform a rigorous, single-metric-based multi-dimensional evaluation of Virtual Try-On (VTON) results. You possess a keen eye for anatomical accuracy, temporal consistency, and physical realism in generative video.

\textbf{Evaluation Task:}
Analyze the provided <Original Video>, <Reference Image>, and <Edited Video>. You must generate a score for each of the following 9 metrics, strictly adhering to the specified Required Assets and Constraints for each.

\textbf{1. Action Trajectory \& Pose Synchronization}
\textbf{Required Assets:} Compare <Original Video> with <Edited Video> (Ignore Reference Image).

\textbf{Metric Definition:} This metric evaluates whether the anatomical movement trajectory and skeletal posture sequence of the model in the <Edited Video> are pixel-perfectly synchronized with the <Original Video>.

\textbf{[CRITICAL - WHAT TO IGNORE]:} This is an Appearance Transfer task. You MUST completely IGNORE differences in:
- Facial identity, makeup, or expressions.
- Clothing style, color, texture, or volume (e.g., a tight shirt changing to a loose dress).
- Added or removed accessories (bags, glasses, jewelry).
DO NOT lower the score if the person or the clothes look completely different. Focus EXCLUSIVELY on the underlying human skeleton and action timing.

\textbf{[SCORING RUBRIC (1-10)]:}
[9-10] Perfect Alignment: The model's body pose, limb placement (hands/feet), head orientation, and movement speed (timing) are pixel-perfectly synchronized across all frames. The anatomical action is identical.
[7-8] Minor Deviations: The overall action is identical, but there are extremely slight spatial offsets (e.g., an arm is positioned a few pixels off) or negligible micro-stutters in timing that do not change the nature of the pose.
[4-6] Noticeable Desynchronization: Obvious spatial deviations (e.g., the head is turned at a visibly different angle, the stance is wider/narrower) OR temporal lagging (e.g., the edited model raises their hand a fraction of a second later than the original).
[1-3] Severe Failure: Completely mismatched poses, missing actions (e.g., the original model walks but the edited model stands still), or severe frame freezing.

\textbf{2. Non-Target Preservation \& Background Integrity}
\textbf{Required Assets:} Compare <Original Video> with <Edited Video> based on the <Reference Image>.

\textbf{Metric Definition:} This metric evaluates whether the AI strictly preserved EVERYTHING from the <Original Video> that was NOT explicitly targeted for change by the <Reference Image>. The Reference Image acts as your "exemption list"—only items in the reference are allowed to change.

\textbf{[CRITICAL - WHAT TO IGNORE (The "Safe Zone")]:} You MUST completely IGNORE changes to the specific garments, accessories, or facial identity that are explicitly presented in the <Reference Image>. Do NOT penalize the video for successfully transferring these target elements.

\textbf{[CRITICAL - WHAT TO PENALIZE (The "Danger Zone")]:} Everything else MUST remain pixel-perfectly identical to the <Original Video>. You must heavily penalize any of the following unintended changes:
- Background \& Environment: Warping, blurring, or color shifts in walls, furniture, floors, or outdoor scenery.
- Non-Target Garments: If the reference only shows a shirt, the model's original pants/shoes MUST NOT change color, shape, or texture.
- Global Context: Changes to overall video lighting, shadows, or adding/hallucinating random objects in the scene.

\textbf{[SCORING RUBRIC (1-10)]:}
[9-10] Perfect Preservation: The background, environmental lighting, and all non-target elements are flawlessly identical to the original video. Only the target items changed.
[7-8] Minor Deviations: Barely noticeable lighting shifts, extremely slight pixel swimming in the deep background, or negligible variations in non-target areas that do not distract the viewer.
[4-6] Noticeable Alterations: Obvious warping/distortion in the background walls/objects, noticeable color bleeding into non-target clothing (e.g., pants subtly changed color), or missing original shadows.
[1-3] Severe Failure: Completely replaced or destroyed background, heavily corrupted non-target garments, or massive hallucinated objects completely changing the scene context.

\textbf{3. Overall Visual Quality}
\textbf{Required Assets:} Analyze EXCLUSIVELY the <Edited Video> (Do NOT compare with Original Video or Reference Image).

\textbf{Metric Definition:} This metric exclusively evaluates the absolute technical clarity, sharpness, and the presence of AI-generated visual artifacts in the <Edited Video>.

\textbf{[WHAT TO PENALIZE (The "Artifact Zone")]:} You must strictly penalize any of the following intrinsic AI-induced degradations:
- Blurriness \& Low Resolution: The video looks soft, out of focus, or lacks high-frequency details (e.g., muddy textures).
- Unnatural Smoothing: Skin, fabrics, or surfaces look like plastic, overly smoothed, or structurally "melted" (a common AI generation artifact).
- Noise, Pixelation \& Glitches: Unwanted grain, blocky artifacts, color banding, or random visual glitches (e.g., floating pixels or corrupted patches).

\textbf{[SCORING RUBRIC (1-10)]:}
[9-10] Pristine Quality: The video is perfectly sharp, crisp, and completely indistinguishable from real, high-quality, professional camera footage. Zero visible AI artifacts.
[7-8] Minor Degradation: Extremely slight softness, minor noise in deep shadows, or negligible compression artifacts. The video still feels premium and realistic.
[4-6] Noticeable Artifacts: Obvious AI artifacts such as visible pixelation, unnatural plastic-like smoothing on skin/clothes, patchy/melted textures, or noticeable blurriness that degrades the realism.
[1-3] Severe Failure: Heavy distortion, extreme blur, severe image corruption, massive glitches, or looking like a low-quality, unusable generation.

\textbf{4. Target Content Completeness}
\textbf{Required Assets:} Compare <Original Video> with <Edited Video> based on the <Reference Image>.

\textbf{Metric Definition:} This metric evaluates whether the AI accurately generated ALL required elements—and ONLY the required elements—prescribed by the <Reference Image> (e.g., specific garments, bags, accessories, or facial identity).

\textbf{[CRITICAL - THE "PRESENCE vs ABSENCE" RULE]:} You are essentially performing a checklist audit. Focus STRICTLY on whether items are missing or hallucinated.
- The "MUST ADD" Rule: If the Reference Image introduces a specific item (e.g., a shoulder bag, new shoes, a specific necklace, or a different facial identity), it MUST be present in the Edited Video.
- The "NO HALLUCINATION" Rule: If the Reference Image DOES NOT contain an item, the AI MUST NOT hallucinate or arbitrarily add extra items (e.g., adding a random hat which not in the reference image).
- The "KEEP UNCHANGE" Rule: Do Not change the parts that are not mentioned in reference images.

\textbf{[SCORING RUBRIC (1-10)]:}
[9-10] Perfect Completeness: Flawless execution of the checklist. Every single target element from the reference is transferred. Zero missing items, and zero hallucinated (invented) objects.
[7-8] Minor Omissions/Additions: Missing an extremely tiny, negligible detail (e.g., a small ring or a tiny button from the reference) OR adding a barely noticeable, tiny visual artifact that resembles a small accessory.
[4-6] Noticeable Failures: Missing a clear, noticeable target (e.g., the AI changed the clothes but failed to add the bag/necklace from the reference, or failed to update the face when the reference clearly provided one) OR hallucinating a noticeable extra item not present in the reference.
[1-3] Severe Failure: Fundamentally failed to generate the main target (e.g., the main garment was not changed at all), completely ignored the reference image, or hallucinated massive, scene-ruining objects.

\textbf{5. Material \& Detail Fidelity}
\textbf{Required Assets:} Evaluate the <Edited Video> based on the <Reference Image>.

\textbf{Metric Definition:} This metric evaluates how accurately the surface details—specifically colors, fabric textures, complex patterns, and text/logos—of the target items from the <Reference Image> are reconstructed in the <Edited Video>.

\textbf{[CRITICAL - WHAT TO IGNORE (The "Surface Only" Rule)]:} You MUST completely IGNORE whether the item fits well, whether its size/proportions are correct, or whether it inherits original occlusions. Those are evaluated elsewhere. Your ONLY focus is on the surface properties: "Does it look made of the exact same material and print as the reference?"

\textbf{[CRITICAL - THE "TEXT \& LOGO" HIGHLINE]:} Generative AI often struggles with typography. Pay extreme attention to any text, graphics, or brand logos present in the Reference Image. If they appear scrambled, blurred, or turn into illegible gibberish in the Edited Video, you MUST penalize it heavily.

\textbf{[SCORING RUBRIC (1-10)]:}
[9-10] Flawless Fidelity: Colors, fabric properties (e.g., the sheen of silk, the fuzziness of wool), and patterns perfectly match the Reference Image. All text and logos are crisp, readable, and structurally identical.
[7-8] Minor Deviations: Very slight color shifts (e.g., slightly warmer/cooler tone) or minor loss of high-frequency fabric texture detail. Logos/text are slightly soft but still entirely accurate and readable.
[4-6] Noticeable Degradation: Obvious pattern mismatches, incorrect fabric properties (e.g., a leather bag looks like plastic), or scrambled/unreadable text and logos (AI gibberish).
[1-3] Severe Failure: The design, color, or texture is fundamentally wrong. Major patterns are completely lost, or the generated item looks like a completely different piece of clothing/accessory than the reference.

\textbf{6. Geometric Accuracy \& Proportions}
\textbf{Required Assets:} Evaluate the <Edited Video> based on the <Reference Image>.

\textbf{Metric Definition:} This metric evaluates the structural fit, sizing, length proportions, and overall spatial geometry of the generated garments and accessories on the model's body, compared to the intended design in the <Reference Image>.

\textbf{[CRITICAL - WHAT TO IGNORE (The "Structure Only" Rule)]:} You MUST completely IGNORE surface details (colors, logos, textures), whether items are missing, or how they interact with environmental occlusions (e.g., hiding behind a desk). Your ONLY focus is: "Is the clothing shaped correctly (e.g., V-neck vs. crew neck, short vs. long sleeves) and are the accessories reasonably sized relative to the human body?"

\textbf{[CRITICAL - WHAT TO PENALIZE (The "Illogical Fit" Zone)]:} You must penalize any structural or scaling hallucinations, such as:
- Garment Structure Alteration: The AI fundamentally changed the design silhouette (e.g., turning shorts into long pants, or a tight crop top into a baggy sweater).
- Unnatural Sizing/Proportions: An added accessory (like a handbag or glasses) is absurdly massive or comically tiny compared to a normal human scale.
- Anatomical Defiance: The clothing sits on the body in a way that defies human skeletal structure (e.g., shoulders floating unrealistically high, necklines placed on the chest).

\textbf{[SCORING RUBRIC (1-10)]:}
[9-10] Perfect Fit \& Scale: The garment structure (sleeves, collars, hemlines) perfectly matches the Reference Image's intent, and all accessories are scaled logically to the model's body.
[7-8] Minor Deviations: Slightly unnatural drape, negligible sizing issues (e.g., a collar slightly wider than intended, or a bag marginally larger than expected), but overall structurally sound.
[4-6] Noticeable Structural Errors: Incorrectly altered garment structure (e.g., short sleeves generated as long sleeves), highly unnatural tightness/bagginess, or noticeably illogical accessory/shoe sizes.
[1-3] Severe Failure: The scale is completely absurd (e.g., giant buttons, a purse larger than the torso), or the geometry fundamentally misaligns with the human skeleton.

\textbf{7. Spatiotemporal Stability}
\textbf{Required Assets:} Analyze EXCLUSIVELY the <Edited Video> (Do NOT compare with Original Video or Reference Image).

\textbf{Metric Definition:} This metric evaluates the temporal consistency of the generated content's pixels and boundaries during motion. The generated elements should appear stable and firmly "locked" onto the moving model.

\textbf{[CRITICAL - WHAT TO IGNORE]:} IGNORE all static properties like color accuracy, material fidelity, or geometric fit. Your ONLY focus is on DYNAMIC artifacts that appear when the video is played.

\textbf{[CRITICAL - WHAT TO PENALIZE (The "Instability Zone")]:} You must strictly penalize two types of temporal failures:
- Texture Instability: This includes flickering (rapid flashing of colors/patterns), boiling (textures seeming to bubble or swim in place), or logos/graphics that morph and change shape frame-to-frame.
- Contour Instability: This includes drifting (garment boundaries sliding away from the body), lagging (garment failing to keep up with fast movements), or edge jittering (the silhouette of the clothes shaking unnaturally).

\textbf{[SCORING RUBRIC (1-10)]:}
[9-10] Flawlessly Stable: Generated content is perfectly locked to the model. Textures and boundaries are completely stable with zero flickering or drifting, even during fast motion.
[7-8] Minor Instability: Barely perceptible micro-jitter on garment edges during very fast movements OR extremely slight texture swimming in shadowed areas. Not distracting.
[4-6] Noticeable Instability: Obvious flickering of patterns/logos that is distracting to the viewer. Garment boundaries clearly detach, slide, or lag behind the body's movement.
[1-3] Severe Failure: Unwatchable, severe flashing or boiling textures across the entire garment. The clothing or accessories completely detach and float in the air, or boundaries violently snap back and forth.

\textbf{8. Anatomical Integrity \& Anti-Clipping}
\textbf{Required Assets:} Analyze EXCLUSIVELY the <Edited Video> (Do NOT compare with Original Video or Reference Image).

\textbf{Metric Definition:} This metric evaluates if the human anatomy remains biologically plausible and structurally intact, and if the generated garments respect physical boundaries without "clipping" through the body.

\textbf{[CRITICAL - WHAT TO IGNORE]:} IGNORE all other quality aspects (color, texture, fit, stability). Your ONLY focus is on the structural integrity of the human form and the physical separation between cloth and skin.

\textbf{[CRITICAL - THE "UNCANNY VALLEY" HIGHLINE]:} This is the most critical metric for realism. Any failure here is a severe error. You must be extremely strict in penalizing:
- Anatomical Corruption: Any malformation of the human body. This includes, but is not limited to: fused, extra, or missing fingers; broken or unnaturally bent limbs (e.g., "rubber hose" arms); distorted necks or torsos.
- Clipping \& Intersection: Any instance where the garment or an accessory illogically merges with or passes through the model's body. This includes fabric sinking into skin, or bag straps slicing through an arm.

\textbf{[SCORING RUBRIC (1-10)]:}
[9-10] Flawless Integrity: Perfect human anatomy across all frames. Boundaries between clothing and skin are clean and distinct. Zero clipping.
[7-8] Minor Boundary Issues: Anatomy is perfect. However, there might be slightly blurred or aliased edges between the garment and skin, but no actual physical intersection.
[4-6] Noticeable Failures: Obvious clipping, where the garment visibly sinks into the body, OR slight but noticeable anatomical distortions, especially around hands and limbs when interacting with the new clothing.
[1-3] Severe Failure \& Anatomical Horror: Horrific and severe anatomical collapse (e.g., mangled hands, missing limbs, completely broken skeletal structure). Violent clipping where clothing passes through large parts of the body. This makes the video unusable and disturbing.

\textbf{9. Physical Dynamics \& Realism}
\textbf{Required Assets:} Analyze EXCLUSIVELY the <Edited Video> (Do NOT compare with Original Video or Reference Image).

\textbf{Metric Definition:} This metric evaluates if the generated garments and accessories behave according to the laws of physics, exhibiting realistic weight, drape, and motion-induced deformations (wrinkles, folds, and sway).

\textbf{[CRITICAL - WHAT TO IGNORE]:} IGNORE all other quality aspects (color, fit, anatomical errors, stability). Your ONLY focus is on the behavior of the cloth and accessories in motion: "Does it move and deform like real fabric, or does it feel stiff and fake?"

\textbf{[CRITICAL - WHAT TO PENALIZE (The "Unnatural Behavior" Zone)]:} You must strictly penalize any violations of physical plausibility:
- "Cardboard Effect" / Rigidity: The clothing lacks natural folds and wrinkles when the model bends or twists, appearing unnaturally stiff like a piece of cardboard or plastic.
- Gravity Defiance: Skirts, sleeves, or bag straps do not hang or swing naturally according to gravity and momentum. They might appear weightless or float artificially.
- Lack of Secondary Motion: The fabric shows no subtle secondary motion, such as small ripples or bounces, when the model stops moving. It feels lifeless.

\textbf{[SCORING RUBRIC (1-10)]:}
[9-10] Flawlessly Realistic: The fabric and accessories exhibit natural weight, drape, and fluid dynamics. Wrinkles, folds, and sways form and resolve organically in response to every movement, perfectly mimicking real-world cloth behavior.
[7-8] Minor Stiffness: The overall motion is plausible, but the fabric feels slightly stiff, lacking some of the finer, subtle secondary wrinkles and bounces. Still believable.
[4-6] Noticeably Unnatural: The fabric exhibits a clear "Cardboard Effect," looking rigid and failing to produce appropriate wrinkles during movement. Skirts or accessories swing in an obviously artificial or "game-like" manner.
[1-3] Severe Failure: A complete breakdown of physics. The clothing is completely static and pasted onto the moving model, or it moves in a bizarre, gravity-defying way that is visually jarring and completely fake.

\textbf{OUTPUT FORMAT:}
Provide your evaluation in the following JSON structure. Output strictly raw, parseable JSON text ONLY. Do NOT wrap the JSON in markdown formatting and do not include any conversational text.

\{
"action\_trajectory\_synchronization": \{
"score": <1-10>,
"reasoning": "[Observation]...[Issues]...[Final Evaluation]..."
\},
"non\_target\_preservation": \{
"score": <1-10>,
"reasoning": "[Observation]...[Issues]...[Final Evaluation]..."
\},
"overall\_visual\_quality": \{
"score": <1-10>,
"reasoning": "[Observation]...[Issues]...[Final Evaluation]..."
\},
"target\_content\_completeness": \{
"score": <1-10>,
"reasoning": "[Observation]...[Issues]...[Final Evaluation]..."
\},
"material\_and\_detail\_fidelity": \{
"score": <1-10>,
"reasoning": "[Observation]...[Issues]...[Final Evaluation]..."
\},
"geometric\_accuracy\_and\_proportions": \{
"score": <1-10>,
"reasoning": "[Observation]...[Issues]...[Final Evaluation]..."
\},
"spatiotemporal\_stability": \{
"score": <1-10>,
"reasoning": "[Observation]...[Issues]...[Final Evaluation]..."
\},
"anatomical\_integrity\_and\_anti\_clipping": \{
"score": <1-10>,
"reasoning": "[Observation]...[Issues]...[Final Evaluation]..."
\},
"physical\_dynamics\_and\_realism": \{
"score": <1-10>,
"reasoning": "[Observation]...[Issues]...[Final Evaluation]..."
\}
\}
\end{tcolorbox}

\subsection{Fine-Tuning Configuration}
\label{sec:finetune_config}
As stated in the implementation details of the main paper, we build OmniTryOn on the pretrained Video-As-Prompt~\cite{bian2025video} backbone (fine-tuned from Wan2.1-I2V-14B~\cite{wan2025wan}) and keep the VAE frozen during training. To further reduce the parameter cost, preserve the priors of the backbone, and retain its generalization ability, we fine-tune only a subset of DiT blocks that amounts to about one quarter of the DiT parameters, while leaving the remaining blocks frozen.

\subsection{RoPE Ablation Variants Configuration}
\label{sec:rope_variants}
We provide the explicit coordinate assignments used by the three biased-RoPE variants compared in the main paper. Let $(t, x, y)$ denote the 3D RoPE coordinates of the target video, with $t\in[0, f-1]$, $x\in[0, h-1]$, and $y\in[0, w-1]$, and let $(t', x', y')$ denote those of the reference video. The three variants and STC-RoPE are configured as:
\begin{itemize}
    \item \textbf{Temporal-biased RoPE}~\cite{bian2025video}: the reference video is placed before the target along the temporal axis, i.e., $t' \in [-f, -1]$, $x' = x$, $y' = y$, so that its tokens occupy a disjoint temporal interval from the target tokens.
    \item \textbf{Spatial-biased RoPE}~\cite{tan2025ominicontrol, wu2025less}: the reference video is shifted along the spatial axes, i.e., $t' = t$, $x' = x + h$, $y' = y + w$, so that its tokens occupy a disjoint spatial region from the target tokens.
    \item \textbf{Spatiotemporal-biased RoPE}: a combination of the above two, i.e., $t' \in [-f, -1]$, $x' = x + h$, $y' = y + w$.
    \item \textbf{STC-RoPE (ours)}: identical coordinates, i.e., $t' = t$, $x' = x$, $y' = y$.
\end{itemize}
All variants share the same training data, backbone, and optimization schedule, and are evaluated after Stage 1 of GTO.

\section{Human Study for VQA Protocol Reliability}
\label{sec:user_study}
As noted in the main paper, we conduct a pairwise user study on the TryAny-Bench test set to verify that the VQA-based evaluation protocol tracks human preference. For each of the five baselines (CatV$^2$TON~\cite{chong2025catv2ton}, MagicTryOn~\cite{li2025magictryon}, ViViD~\cite{fang2024vivid}, VACE~\cite{jiang2025vace}, and VAP~\cite{bian2025video}), we form 217 non-cherry-picked pairs, each containing the OmniTryOn result and the baseline result on the same test sample, and ask annotators to select the preferred result or a tie.

\begin{figure}[htbp]
    \centering
    \includegraphics[width=\linewidth]{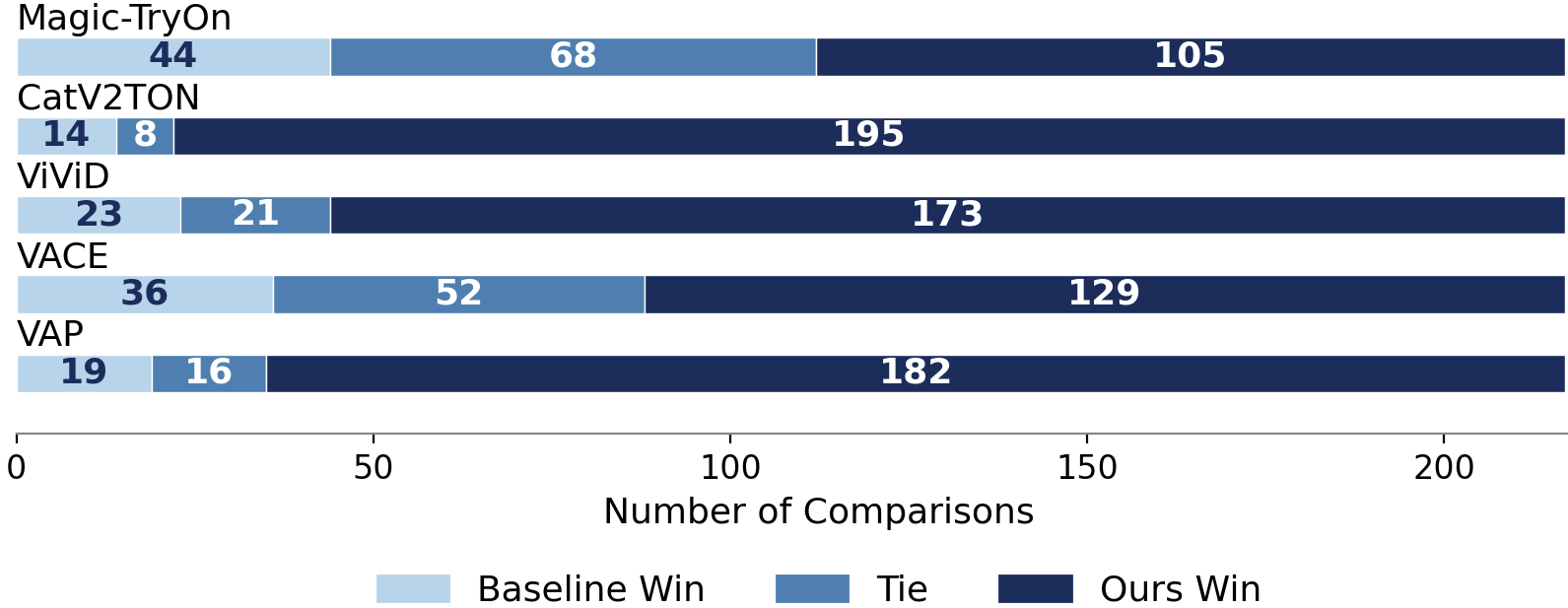}
    \caption{Pairwise user study on 217 comparisons per baseline. Bars report the number of samples where the OmniTryOn result is preferred, tied, or the baseline result is preferred.}
    \label{fig:user_study}
\end{figure}

As illustrated in Fig.~\ref{fig:user_study}, OmniTryOn is preferred over every baseline. Excluding ties, human annotators favor OmniTryOn in $70.5\%$ to $93.3\%$ of the comparisons, and the induced pairwise ranking of baselines is identical to that induced by the VQA-based evaluation. This alignment supports the reliability of the VQA-based protocol as a scalable proxy for human judgment in the Try-On Anything task.

\begin{table*}[t]
\centering
\resizebox{\textwidth}{!}{
\begin{tabular}{llcccccc}
\toprule
\textbf{Dimension} & \textbf{Metric} & \textbf{CatV2TON~\cite{chong2025catv2ton}} & \textbf{MagicTryOn~\cite{li2025magictryon}} & \textbf{VACE~\cite{jiang2025vace}} & \textbf{VAP~\cite{bian2025video}} & \textbf{ViViD~\cite{fang2024vivid}} & \textbf{Ours} \\ 
\midrule
\multirow{3}{*}{\textit{Video Quality}} 
& Visual Fidelity (VF)        & 2.49 & 7.50 & 8.00 & 7.96 & 6.11 & \textbf{8.80} \\
& Action Synchronization (AS) & 9.07 & \textbf{10.00} & 9.96 & 5.66 & 9.88 & 9.92 \\
& Environment Stability (ES)  & 5.35 & 9.18 & 9.09 & 3.61 & 8.23 & \textbf{9.26} \\ 
\midrule
\multirow{3}{*}{\textit{Try-On Stability}} 
& Object Integrity (OI)       & 4.30 & 4.69 & 5.06 & 6.98 & 4.56 & \textbf{9.25} \\
& Material Fidelity (MF)      & 2.87 & 6.44 & 6.99 & 7.65 & 5.71 & \textbf{8.86} \\
& Scale Naturalness (SN)      & 4.82 & 8.29 & 8.24 & 7.82 & 7.04 & \textbf{9.55} \\ 
\midrule
\multirow{3}{*}{\textit{Physical Realism}} 
& Temporal Stability (TS)     & 2.83 & 7.77 & 7.78 & 7.99 & 5.88 & \textbf{8.90} \\
& Anatomical Integrity (AI)   & 4.90 & 9.09 & 9.07 & 8.19 & 7.68 & \textbf{9.59} \\
& Dynamic Plausibility (DP)   & 2.88 & 7.65 & 8.05 & 7.15 & 6.31 & \textbf{8.94} \\ 
\midrule
\multicolumn{2}{c}{\textbf{Average Score (9D)}} & 4.39 & 7.85 & 8.03 & 7.00 & 6.82 & \textbf{9.23} \\ 
\bottomrule
\end{tabular}
}
\caption{Detailed quantitative results on the TryAny-Bench utilizing the multidimensional VQA evaluation protocol. The metrics are grouped into three primary evaluation dimensions. The best results are highlighted in \textbf{bold}.}
\label{tab:vqa_detailed_results}
\end{table*}

\section{Detailed Quantitative Results of TryAny-Bench}
\label{sec: Detailed Quantitative Results of TryAny-Bench}

In the main paper, we provided a multi-dimensional radar chart to intuitively illustrate the superior capabilities of OmniTryOn across various evaluation axes. To further ensure data transparency and facilitate future comparisons, we present the exhaustive quantitative results of our VQA-based evaluation protocol in Table~\ref{tab:vqa_detailed_results}. 

The metrics are systematically categorized into three primary dimensions: \textit{Video Quality}, \textit{Try-On Stability}, and \textit{Physical Realism}. As shown in the table, OmniTryOn consistently outperforms existing state-of-the-art baselines by a significant margin across almost all sub-metrics. Notably, in critical areas such as Object Integrity (OI) and Scale Naturalness (SN), which directly reflect the model's ability to handle multi-object customization and geometric accuracy, OmniTryOn achieves striking improvements.

\section{Analysis of the Dataset Construction Pipeline}
\label{sec:pipeline_analysis}
The main paper introduces the TryAny-Bench construction pipeline. Because it composes several generative components, we further analyze how quality is preserved end-to-end and how the pipeline is designed to strengthen the generalization ability of the trained model.

\noindent\textbf{Complementary Teachers.} We combine two teachers with complementary strengths. MagicTryOn~\cite{li2025magictryon} is a garment-specialized VVT model that replaces upper and lower garments with high fidelity but tends to leave residual artifacts around masked regions and does not support non-garment objects. CAPYBARA~\cite{capybara2026rao} is a general video editing model that modifies the remaining wearable objects (e.g., handbags and shoes) and, in the same pass, inpaints the residual artifacts left by MagicTryOn. This composition simultaneously supports multi-object modification and restores spatiotemporal consistency, so that the resulting reference videos preserve the human motion and background dynamics of the raw target videos.

\noindent\textbf{Anti-Leakage Augmentation.} When extracting wearable objects with Gemini 3.1 Flash Image~\cite{gemini31flashimage2026}, the extraction prompt perturbs the scale and orientation of each object relative to its appearance in the target video. This augmentation prevents spatial leakage: the model cannot rely on a trivial copy-paste mapping between the wearable object image and the target video, and is instead forced to learn scale- and orientation-invariant transfer.

\noindent\textbf{Manual Quality Control.} After the automated pipeline, we manually inspect every generated pair and discard samples with visible teacher-inherited artifacts. The final dataset contains 1,460 paired samples for training and evaluation.

\section{Additional Visualization Results}
\label{sec:additional_visualization}
As briefly discussed in the qualitative comparison of the main paper, we further validate OmniTryOn under two challenging settings that are complementary to the standard 49-frame in-domain evaluation.

\subsection{In-the-Wild Generalization}
\label{sec:in_the_wild}
TryAny-Bench is built from e-commerce videos, so we additionally examine OmniTryOn on in-the-wild footage outside this training distribution. The top row of Fig.~\ref{fig:in_the_wild} shows one such example, in which the person walks under strong directional lighting in front of an outdoor gate. OmniTryOn transfers the target wearable object onto the person while preserving human motion, camera trajectory, and background structures. We observe a mild lighting mismatch between the transferred wearable and the surrounding scene as the main failure mode, which we attribute to the domain gap between the e-commerce training videos and outdoor illumination.

\subsection{Long-Sequence and Extreme-Motion Stability}
\label{sec:long_sequence}
\begin{figure}[htbp]
    \centering
    \includegraphics[width=\linewidth]{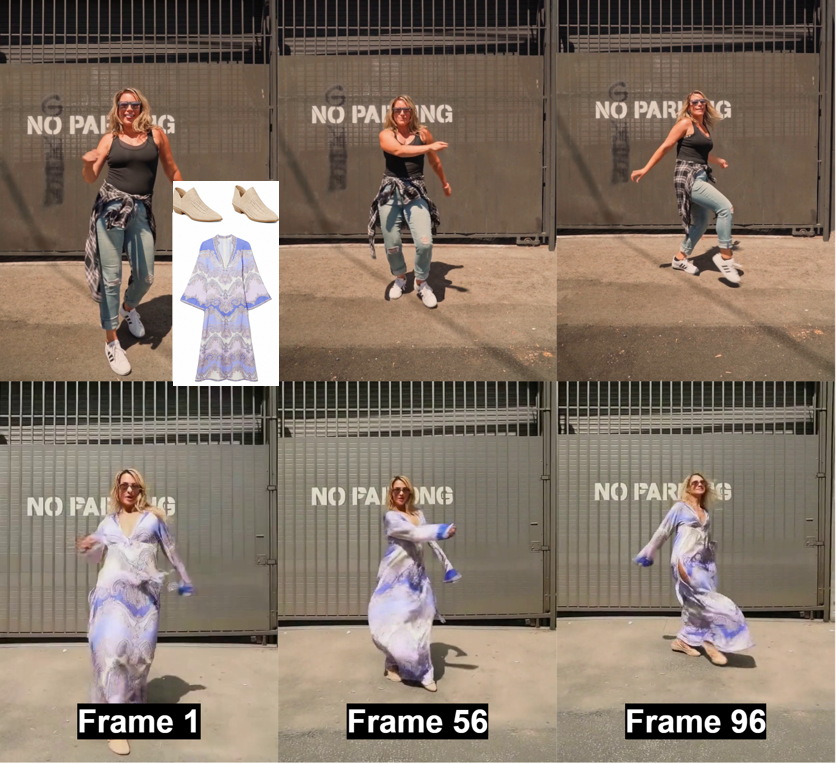}
    \caption{Additional visualization results of OmniTryOn. Top: in-the-wild transfer under strong directional lighting. Bottom: Frame 1, Frame 56, and Frame 96 of a 96-frame sequence with fast motion and heavy occlusion.}
    \label{fig:in_the_wild}
\end{figure}

The bottom row of Fig.~\ref{fig:in_the_wild} shows Frame 1, Frame 56, and Frame 96 of a 96-frame clip in which the person walks with fast body rotation and heavy self-occlusion. The wearable object retains its high-frequency textures across the sequence, and human motion together with the background remains stable, indicating that the First Frame Wearable Cache generalizes to sequences much longer than the 49-frame training length.

\subsection{More Visualization Results}
\label{sec: More Visualization Results}
To further substantiate the superior synthesis capabilities of OmniTryOn, we provide extended visualization results in this section. Constrained by the strict page limits of the main manuscript, we showcase additional high-fidelity video virtual try-on examples here. These supplementary visualizations encompass a broader spectrum of complex wearable combinations, ranging from diverse garments to various accessories such as handbags and footwear, across challenging human motions and background environments. As illustrated in Fig.~\ref{fig1}-Fig.~\ref{fig3}, OmniTryOn consistently exhibits exceptional robustness in achieving seamless multi-object customization while strictly preserving the original spatiotemporal dynamics and physical realism without relying on explicit external priors.








\end{document}